\documentclass[sn-aps]{sn-jnl}

\jyear{2023}%

\theoremstyle{thmstyleone}%
\theoremstyle{thmstyletwo}%

\theoremstyle{thmstylethree}%

\usepackage{color,array,amsthm}
\usepackage{graphicx}
\usepackage{multirow}
\usepackage{verbatim} 
\usepackage{natbib}
\usepackage{algpseudocode}
\usepackage[autostyle]{csquotes}

\algnewcommand\INPUT{\item[\textbf{Input:}]}%
\algnewcommand\OUTPUT{\item[\textbf{Output:}]}%
\algnewcommand\RETURN{\State \algorithmicreturn}
\usepackage{algorithm}
\usepackage{algpseudocode}
\usepackage{algorithmicx}
\usepackage{caption}
\usepackage{subcaption}
\usepackage[font=small,labelfont=bf,margin=\parindent,tableposition=top]{caption}
\usepackage{blindtext}
\usepackage{amsmath}

\begin{document}

\title[ ]
{\textbf{Cooperative Multi-Agent Planning Framework for Fuel Constrained UAV-UGV Routing Problem}}


\author*[1]{\fnm{Md Safwan} \sur{Mondal}}\email{mdsafwanmondal@gmail.com}
\author[1]{\fnm{Subramanian} \sur{Ramasamy}}\email{sramas21@uic.edu}
\author[3]{\fnm{James D.} \sur{Humann} \email{james.d.humann.civ@army.mil}}
\author[2]{\fnm{Jean-Paul F.} \sur{Reddinger} \email{jean-paul.f.reddinger.civ@army.mil}}
\author[2]{\fnm{James M.} \sur{Dotterweich} \email{james.m.dotterweich.civ@army.mil}}

\author[2]{\fnm{Marshal A.} \sur{Childers} \email{marshal.a.childers.civ@army.mil}}
\author[1]{\fnm{Pranav A.} \sur{Bhounsule} \email{pranav@uic.edu}}

\affil*[1]{\orgdiv{Department of Mechanical and Industrial Engineering}, \orgname{University of Illinois Chicago}, \orgaddress{\city{Chicago}, \postcode{60607}, \state{ IL}, \country{USA}}}

\affil[2]{\orgname{DEVCOM Army Research Laboratory}, \orgaddress{\street{Aberdeen Proving
Grounds}, \city{Aberdeen}, \postcode{21005}, \state{MD}, \country{USA}}}

\affil[3]{\orgname{DEVCOM Army Research Laboratory}, \orgaddress{ \city{Los Angeles}, \postcode{90094}, \state{CA}, \country{USA}}}


\abstract{ Unmanned Aerial Vehicles (UAVs), although adept at aerial surveillance, are often constrained by limited battery capacity. By refueling on slow-moving Unmanned Ground Vehicles (UGVs), their operational endurance can be significantly enhanced. This paper explores the computationally complex problem of cooperative UAV-UGV routing for vast area surveillance within the speed and fuel constraints, presenting a sequential multi-agent planning framework for achieving feasible and optimally satisfactory solutions. By considering the UAV fuel limits and utilizing a minimum set cover algorithm, we determine UGV refueling stops, which in turn facilitate UGV route planning at the first step and through a task allocation technique and energy constrained vehicle routing problem modeling with time windows (E-VRPTW) we achieve the UAV route at the second step of the framework. The effectiveness of our multi-agent strategy is demonstrated through the implementation on 30 different task scenarios across 3 different scales. This work offers significant insight into the collaborative advantages of UAV-UGV systems and introduces heuristic approaches to bypass computational challenges and swiftly reach high-quality solutions.}


\maketitle
\small

\section{Introduction}
Over the last decade, unmanned aerial vehicles (UAVs) and unmanned ground vehicles (UGVs) have become increasingly prevalent in a variety of sectors. These applications span from intelligence gathering, surveillance, and reconnaissance tasks \cite{stolfi2021uav, wu2020cooperative,liu2019cooperative}, to search and rescue tasks \cite{li2016hybrid} and even agricultural activities \cite{tokekar2016sensor}. Due to their cost-effectiveness, ease of control, and high maneuverability UAVs are perfect for rapidly scanning or surveying any terrain. However, their usage is primarily suited for small-scale projects due to their limited battery lifespan and small payload capacity.

In contrast, UGVs, equipped with a larger cargo hold and extended battery life, can withstand lengthier task durations. Nevertheless, their efficacy is often compromised by obstacles such as challenging ground terrain, limited visibility, and slower movement speed, which frequently results in incomplete task accomplishment.

To counteract these individual drawbacks, a cooperative routing strategy involving both UAVs and UGVs can be employed. This approach enhances operation coverage speed and endurance. For instance, in an extensive surveillance process, UAVs can reach a set of distant assignment points while being periodically refueled by the UGV, which is acting as a mobile refueling depot. Simultaneously, the UGV can cover assignment points along the road network, thus reducing the UAVs' workload and ensuring the operation's swift completion.

This collaborative approach between UAV and UGV can be modeled as a cooperative vehicle routing problem. In this study, we put forth a framework aimed at efficiently finding the optimal solution to a UAV-UGV cooperative vehicle routing problem that takes into account UAV's fuel constraints.

\section{Related works}
There has been significant research conducted on the cooperative routing of fuel-constrained UAVs with ground vehicles. The routing issue of multiple fuel-constrained UAVs with several static recharging depots had been explored by Levy et al.\cite{levy2014heuristics}. They employed quick variable neighborhood descent (VND) and variable neighborhood search (VNS) heuristics to identify good feasible solutions for large instances. Sundar et al. \cite{sundar2016formulations} further developed a mixed-integer linear programming model (MILP) for the same problem, which was solved using a standard MILP solver. In contrast to fixed charging stations, Maini et al. \cite{maini2015cooperation} addressed a cooperative routing problem involving a single UAV-UGV system, where the UGV has the ability to recharge the UAV while in transit on a road. They proposed a greedy heuristic for determining the meeting points for recharging along the UGV route and later used a MILP model to solve both UAV-UGV routes. Manyam et al. \cite{manyam2019cooperative} examined the cooperative routing of an air and ground vehicle team considering communication constraints. They framed the problem as a mixed-integer linear program and developed a branch-and-cut algorithm to solve the problem to optimality.

Several researchers have delved deeper into the UAV-UGV cooperative vehicle routing problem, exploring it in a tiered, two-echelon manner \cite{li2021ground}. For instance, Luo et al. \cite{luo2017two} introduced a binary integer programming model, supplemented by two heuristics, to tackle this two-echelon cooperative routing challenge. In a related context, Liu et al. \cite{liu2020two} devised a two-stage, route-focused framework for a parcel delivery system that utilized a truck and a drone. This framework aimed to optimize both the primary route of the truck and the associated aerial routes of the drone. To swiftly generate a feasible solution, they developed a hybrid heuristic, which integrated the strategies of nearest neighbor and cost saving. In our previous works \cite{ramasamy2021cooperative,ramasamy2022coordinated, mondal2023optimizing}, we studied a hierarchical, bi-level optimization framework for the cooperative routing of multiple fuel-limited UAVs and a single UGV. The outer level of this framework employed K-means clustering to determine UGV visit points. These points were then connected using a Traveling Salesman Problem (TSP) approach to establish the UGV route. On the inner level, using the determined UGV path, we formulated and solved a vehicle routing problem that took into account capacity constraints, time windows, and dropped visits for the UAV. Further expanding on this work, we demonstrated that optimizing heuristic parameters using Genetic Algorithm (GA) and Bayesian Optimization (BO) methods could lead to substantial improvements in the solution quality \cite{ramasamy2022heterogenous, ramasamy2023solving}. 

Given the intricacy of this problem, exact methods of solving this combinatorial optimization problem or generalizing a solution framework for diverse scenarios pose significant challenges. In this research endeavor, we propose a generalized multi-agent cooperative framework for addressing this fuel-constrained UAV-UGV cooperative routing issue. The key contribution of our study is the creation of heuristics aimed at facilitating a rapid solution to the two-echelon UAV-UGV routing problem, considering fuel and speed constraints. To this end, our novel contributions include the following:
\begin{enumerate}
    
\item  The proposed comprehensive framework utilizes sequential optimization with a task allocation technique. Coupled with the constrained programming-based formulations, it can provide an effective solution for fuel constrained UAV-UGV cooperative routing problems in a quick time.

\item  A task allocation technique based on the minimum set cover algorithm is proposed, which breaks down the entire problem into smaller subproblems, leading to a substantial simplification of the problem-solving process.

\item Our formulation of a constraint programming-based vehicle routing problem accommodates time windows, and fuel constraints, thereby enabling swift solutions for each subproblem.

\item  We present extensive computational results on different kinds of scenarios to affirm the effectiveness and robustness of our proposed framework. This underscores the practicality of our framework in a diverse set of real-world applications.

\end{enumerate}


The rest of the article is structured as follows. Section 3 presents the
problem statement, section 4 illustrates the framework methodology and solution heuristics. Section 5 introduces the experiments of different random instances in three different scales and shows the results part, offering a concrete view of our findings. The results are analyzed in section 6 and finally, section 7 presents the conclusion and outlines future works.

\section{Problem Description}
\begin{figure}[htbp]
\centering
\includegraphics[ scale=0.5]{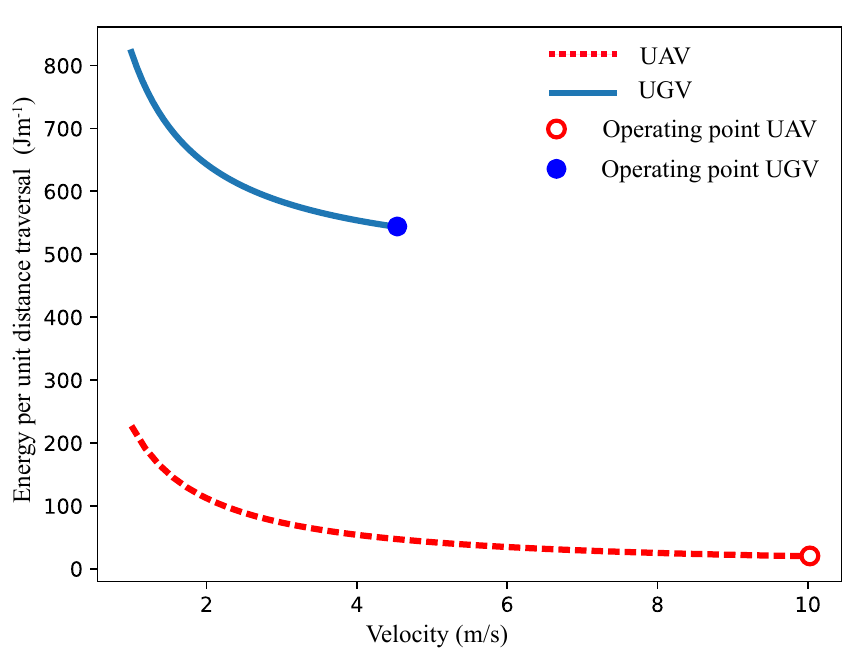}
\caption{Energy consumption per unit distance traversal of UAV \& UGV}
\label{Power_curve}
\end{figure}

\begin{figure*}[t]
\centering
\begin{subfigure}[]{0.3\textwidth}
         \includegraphics[ scale=0.4]{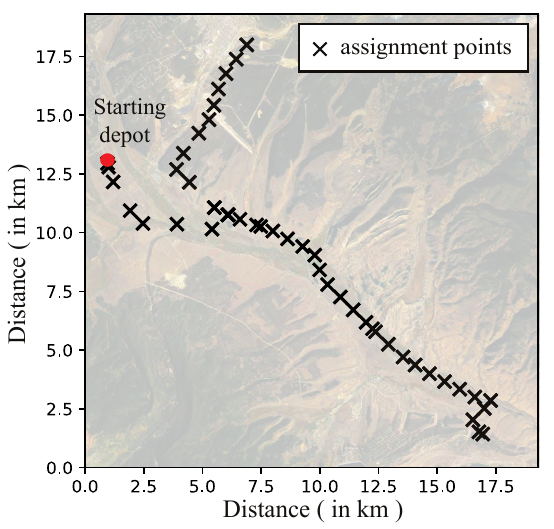 }
         \caption{Given scenario}  \label{given scenario}
\end{subfigure}
\hspace*{2mm}
\begin{subfigure}[]{0.3\textwidth}
         \includegraphics[ scale=0.4]{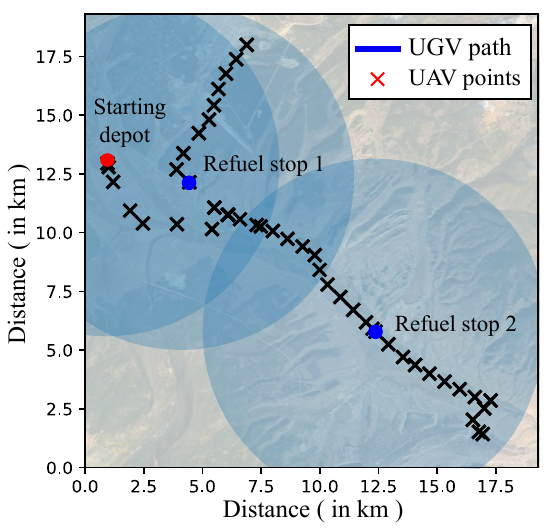}
         \caption{Refuel stop locations}
         \label{entire problem}
\end{subfigure}

\vspace{0.001mm}
\begin{subfigure}[]{0.3\textwidth}
         \includegraphics[ scale=0.4]{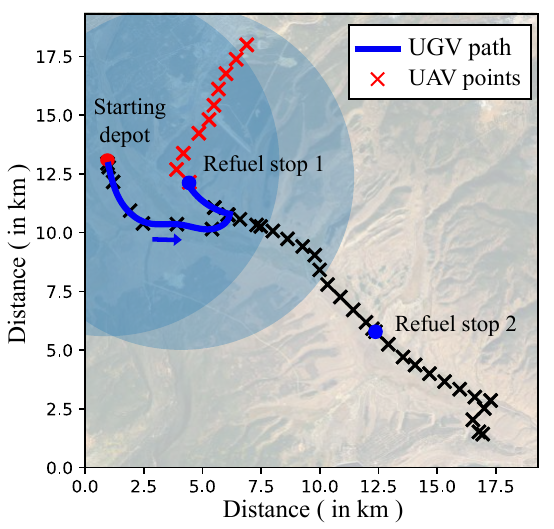}
         \caption{ Subproblem 1}
         \label{subproblem1}
\end{subfigure}
\hspace*{2mm}    
\begin{subfigure}[]{0.3\textwidth}
         \includegraphics[ scale=0.4]{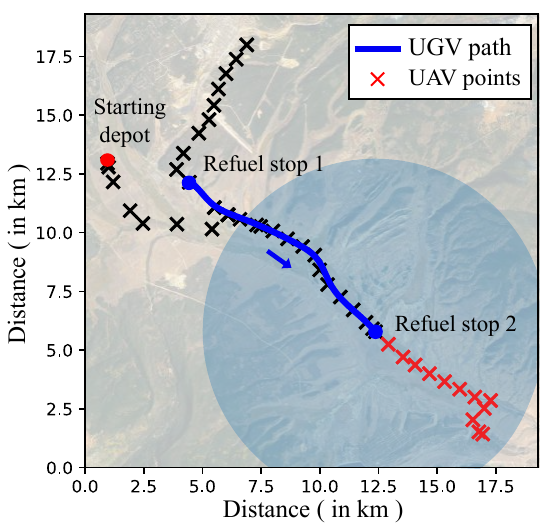}
         \caption{Subproblem 2}
         \label{subproblem2}
\end{subfigure}

\caption{  a) Given scenario with assignment points and starting depot b) Refuel stops in UGV route obtained from minimum set cover algorithm; the blue circles are indicating the radial coverage of the UAV c) Subproblem 1 with allocated UAV assignment points, here UGV travels between starting depot and refuel stop 1  d) Subproblem 2 with allocated UAV assignment points, here UGV travels between refuel stop 1 and refuel stop 2. }
\label{Divison of subproblem}
\end{figure*}

The problem objective is to configure an optimal cooperative route for a team comprising a UAV and UGV to visit a set of $n$ assignment points $ \mathcal{M}_n =\{m_0,m_1,...,m_n\}$ in a euclidean space (see figure \ref{given scenario}). The UAV $ { A \equiv (v^{a},F^a, \mathcal{P}^a) } $ and the UGV $ { G \equiv (v^{g},F^g, \mathcal{P}^g) } $  have heterogeneous vehicle characteristics; with the UAV having a higher velocity, i.e. $v^a > v^g$ but lower fuel capacity than that of the UGV, i.e. $ F^a < F^g $. They also differ in their power consumption profiles (Eq. \ref{UAV-power}, Eq. \ref{UGV-power}), with the UAV demonstrating greater energy efficiency per unit distance traversal when operating at standard speeds (see figure \ref{Power_curve}). The assignment points can be visited by a free flyover of the UAV, $\tau^a$ or a visit via the UGV's road network, $\tau^g$. The cost of travel between a pair of assignment points is equal to the time of traversal between them $t_{ij} = t_j - t_i$. Both the UAV and UGV commence their journeys from the same starting depot and return to it upon completion. The total task duration is the time span from when the first vehicle departs the depot until the last one returns.

\begin{equation}
    \mathcal{P}^a = 0.0461{v^a}^3-0.5834{v^a}^2-1.8761v^a+229.6
    \label{UAV-power}
\end{equation}

\begin{equation}
    \mathcal{P}^g = 464.8v^g+356.3
    \label{UGV-power}
\end{equation}

Due to having a limited battery capacity, the UAV has to get recharged periodically from the UGV which acts as a mobile recharging depot beside visiting the assignment points. The recharging time of the UAV at the UGV is not instantaneous, it depends upon the amount of fuel present in the UAV. Since the fuel capacity of the UGV is significantly larger compared to the UAV, it is assumed to be infinite to simplify the problem.

With all these above configurations, we have to find the time optimal cooperative route, $\tau = \tau^a \cup \tau^g$ between the UAV and UGV for visiting all the assignment points at least once; given the UAV will never run out of fuel. A typical sequence for this cooperative route could be as follows: Both the UAV and UGV commence their journey from the starting depot and visit several assignment points. As they proceed, the UGV will reach an appropriate location for the UAV to recharge. After recharging, both vehicles will resume their task, continuing to visit assignment points until they reach the next recharging stop. This pattern will continue until all assignment points have been visited, after which both the UAV and UGV will return to the starting depot to conclude their task. However, for an optimal cooperative route, it is important to figure out:
\begin {enumerate}
\item Suitable refueling stop locations $\mathcal{M}_r = \{m^r_0,m^r_1,...,m^r_n\}$, \textbf{where} the UAV, UGV will rendezvous for recharging.
\item Appropriate time intervals during the task \textbf{when} the UAV, UGV will meet at the refuel stops i.e, i.e, $t^r_i \ \forall\ m^r_i \in \mathcal{M}_r $.

\item Optimal routes for the UAV, $\tau^a$ and UGV, $\tau^g$ based on the determined refuel stop locations $m^r_i$ and time intervals $t^r_i$ to cover the entire assignment scenario in the quickest possible time.
\end {enumerate}

\section{ Methods}

\begin{figure}[htbp]
\centering
\includegraphics[ scale=0.38]{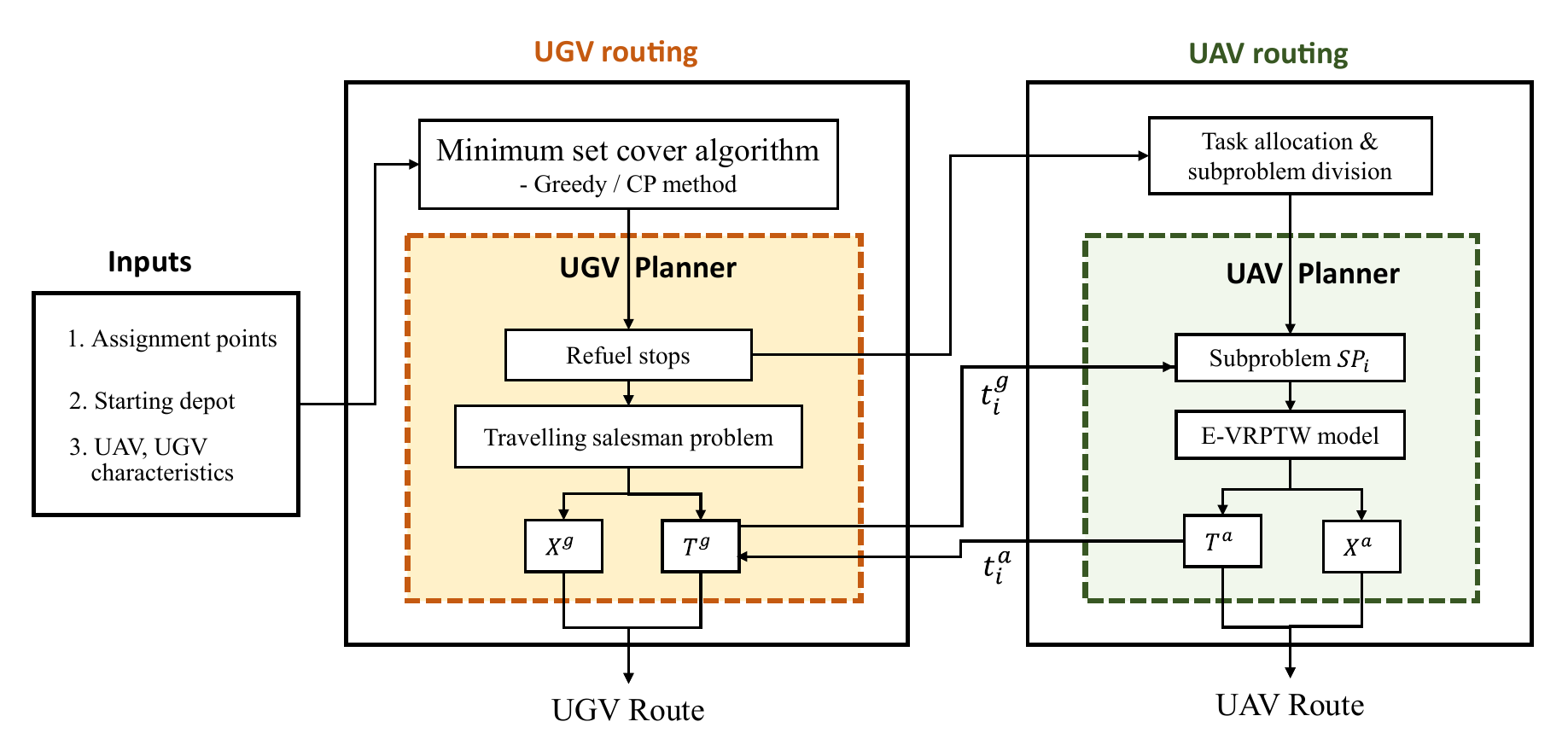}
\caption{proposed framework}
\label{FW}
\end{figure}

We have devised a two-tiered optimization framework (as depicted in figure \ref{FW}) for executing this fuel-constrained cooperative routing between the UAV and UGV. This framework is inspired by the ``UGV First, UAV Second" heuristic approach for UAV-UGV cooperative routing \cite{gao2020commanding}.

At the first stage of this framework, we utilize a \textbf{\textit{UGVPlanner}} to establish the route $\tau^g = (X^g,T^g)$ for the UGV. This route is constructed by identifying appropriate recharging stations $ \mathcal{M}_r $ and formulating the UGV's movement along the road network accordingly. The UGV's navigation is a combination of two-step processes. The initial phase involves movement along waypoints on the road network to cover the assignment points, while the second phase necessitates waiting for the UAV at the recharging stops.

At the second tier of the framework, the \textbf{\textit{UAVPlanner}}   devises the route $\tau^a = (X^a, T^a) $ for the UAV. The formation of this route significantly relies on the route $\tau^g$ created for the UGV at the outer level of the framework. Because of the slower speed of the UGV, the \textbf{\textit{UAVPlanner}} takes into consideration the \textit{availability} time window constraint at the refuel stops. This planning approach effectively divides the entire scenario into a series of manageable subproblems, each of which can be solved by modeling it as an energy-constrained vehicle routing problem with time windows constraint(E-VRPTW).

\subsection{UGV routing}

At the outer level of the proposed framework,  the initial objective is to determine suitable recharging rendezvous locations $\mathcal{M}_r$ for the UAV-UGV system. Subsequently, an optimal route $\tau^g$ is generated for the UGV, taking into account the refuel stop locations $\mathcal{M}_r$ and its operational speed $v^g$ by the \textbf{\textit{UGVPlanner}}. Previous research by Maini et al.\cite{maini2015cooperation, maini2019cooperative}  emphasized the importance of including refueling stops within the UAV's fuel coverage radius to ensure a viable route in fuel-constrained cooperative routing problems. It was also noted that minimizing the number of recharging instances can reduce the time spent on recharging and minimize the detour required in the UAV's route, resulting in a faster cooperative route.  With these aforementioned considerations, we implemented the minimum set cover algorithm (MSC) to find out minimum number of refueling stops and their locations $\mathcal{M}_r$ that cover the entire assignment scenario. The minimum set cover problem has been extensively studied \cite{gens1980complexity,lin2019simple}, and various methods, including greedy approaches \cite{maini2015cooperation, slavik1996tight} have been proposed. However, we came up with an alternative constraint programming formulation for solving the minimum set cover problem in the context of the cooperative routing problem.

\subsubsection{Minimum set cover algorithm}

\begin{enumerate}

    \item  \textbf{Greedy heuristics approach} :

Minimum set cover algorithm is an NP-hard problem, however through the greedy heuristics the complexity of the problem can be reduced significantly. In the context of cooperative routing problem to find optimal refuel stops through greedy algorithm, we start with the assignment points $\mathcal{M}_n$ that needed to be covered, the fuel capacity $F^a$ of the UAV and the starting depot $m_0$ of the scenario. Our goal is to obtain smallest possible subset of $\mathcal{M}_n$ that can act as refueling stops $\mathcal{M}_r$. As shown in algorithm \ref{algorithm:MSC}, greedy algorithm includes the stating depot $m_0$ as the first refueling stop $m^r_0$, then it sequentially adds (line 6) the assignment points $m_i$ which are covering maximum number of other assignment points into the refueling stop set $m^r_i$ until all the points are covered.  

Greedy heuristics can quickly generate optimal result for a minimum set cover problem. But in many situations a minimum set cover problem can have multiple optimal results for a particular scenario; because we are implementing  a bilevel optimization framework it is important to take into account of the other optimal solutions of the outer level algorithm. As it is not possible to acquire all the optimal solutions through greedy heuristics, we used constraint programming method which can generate multiple optimal results (if any) in a quick span of time.

\begin{algorithm}
    \caption{Greedy Minimum Set Cover Algorithm}
    \begin{algorithmic}[1]
        \INPUT assignment points $\mathcal{M}$, UAV fuel limit $F^a$, starting depot $m_0$;
        \OUTPUT Refueling stops $\mathcal{M}_r$;
        \State Initialize $\mathcal{M}_r = \{m^r_0= m_0\}$, Tasks = $\mathcal{T} = \mathcal{M}$;
        \State $C_0 = Covered(m^r_0) = \{m_i: m_i \in \mathcal{T} \text{ and } \|m_i - m^r_0\| < 0.5F^a \}$;
        \State $\mathcal{T} = \mathcal{T} \setminus C_0$;
        \While {$\mathcal{T} \neq \emptyset$}
        \State $m^r_{i_{\text{max}}} = \arg\max\ Covered(m^r_i) $;
        \State $\mathcal{M}_r = \mathcal{M}_r \cup \{m^r_{i_{\text{max}}}\}$;
        \State  $C_{\text{max}} = \{m_i: m_i \in \mathcal{T} \text{ and } \|m_i - m^r_{i_{\text{max}}}\| < 0.5F^a \}$;
        \State $\mathcal{T} = \mathcal{T} \setminus C_{\text{max}}$;
        \EndWhile
    \end{algorithmic}
    \label{algorithm:MSC}
\end{algorithm}

\item \textbf{Constraint programming method :}

To determine the minimum number of refueling stops $\mathcal{M}_r$  required to cover the entire assignment scenario ($\mathcal{M}$), we employ linear integer programming and utilize a constraint programming method (CP method) for solving. The problem is modeled using binary decision variables, $x_j$ (indicating whether an assignment point is chosen as a refueling stop) and $y_{ij}$ (indicating whether an assignment point $m_i$ is assigned to a refueling stop $m^r_j$). The objective function (Eq. \ref{cp_obj}) aims to minimize the total number of refueling stops. Constraint (Eq.\ref{cp_cns}) ensures that each assignment point $m_i$ is assigned at least one refueling stop $m^r_j$. Constraint (Eq. \ref{cp_cns1}) ensures that an assignment point $m_i$ can only be allocated to a refueling stop $m^r_j$ if the refueling stop is selected. Furthermore, constraint (Eq. \ref{cp_cns2}) guarantees that an assignment point $m_i$ is assigned to a refueling stop $m^r_j$ only if the refueling stop falls within the fuel coverage radius of the UAV, allowing for a round trip from the refueling stop.

\begin{equation}
\text{Objective:     }  \min \sum_{m^r_j \in \mathcal{M}_r}x_j 
\label{cp_obj}
\end{equation}

\text{Subject to,} 
\begin{equation}
 \sum_{m^r_j \in \mathcal{M}_r} y_{ij} \geq 1, \; \forall\ m_i \in \mathcal{M}
\label{cp_cns}
\end{equation}

\begin{equation}
y_{ij} \leq x_j, \; \forall\ m_i \in \mathcal{M}\ \text{and} \ \forall\ m^r_j \in \mathcal{M}_r 
\label{cp_cns1}
\end{equation}

\begin{equation}
y_{ij} = 0, \; \text{if} \; d_{ij} > 0.5F^a, \; \forall\ m_i \in \mathcal{M}\ \text{and} \ \forall\ m^r_j \in \mathcal{M}_r
\label{cp_cns2}
\end{equation}

\begin{equation}
y_{ij}, x_j \in \{0,1\} 
\label{cp_cns3}
\end{equation}

We had used Google's OR-Tools\texttrademark \ \ Constraint programming solver ( CP-SAT solver \cite{ORtools}) to solve the above linear integer formulation. It is possible to record all the solutions if there are multiple optimal solutions through the solver. Once, the optimal refuel stops $\mathcal{M}_r$ are obtained from the MSC algorithm, it is sent to the \textit{\textbf{UGVPlanner}} to construct the UGV route $\tau^g = (X^g, T^g)$ based on it.

\end{enumerate}

\subsubsection{UGV Planner}
Upon identifying the refueling stop locations $\mathcal{M}_r$ using the minimum set cover algorithm, the \textbf{\textit{UGVPlanner}} proceeds to map out a feasible UGV route for the overall task through a sequential phase process (see algorithm \ref{alg:UGVPlanner}). Initially, it connects the refueling stops optimally on the road network by solving a simple Travelling Salesman Problem (TSP), what yields the spatial components $X^g \in \tau^g$  of the UGV route , denoting the sequence $x^g_i$ in which the assignment points on the road network will be visited. Next, the planner calculates the temporal components $T^g \in \tau^g$ of the UGV route till the first refuel stop, which details the time instances at which the UGV will visit those assignment points. We operate under the assumption that the UGV will not wait at any assignment point, except at the refueling stops. Therefore, the arrival times at the assignment points are computed based on the UGV's constant operational speed $v^g$ (line 4). This also gives the UGV's arrival time at the refueling stops (line 7), which serve as an $availability$ time window constraint in the \textbf{\textit{UAVPlanner}}. 
Utilizing the UAV's arrival time at the first refuel stop and the recharging time $\mathcal{R}_t$ (contingent on the UAV's fuel consumption level) from the UAV's route in subproblem 1, we can estimate the UGV's waiting time at the first refuel stop (line 9), which is taken into account when computing the temporal component of the UGV route up to the next refuel stop. This process is reiterated until the UGV arrives at the final refuel stop. 

At the end of this process, the temporal components are integrated with their respective spatial components to provide a comprehensive UGV route, outlining the sequence in which the UGV visits the assignment points and their corresponding time instances.

\begin{algorithm}
\caption{UGV Planner}\label{alg:UGVPlanner}
\begin{algorithmic}[1]
\Statex \textbf{Input:} Refuel stops $\mathcal{M}_r \gets MSC $, UGV velocity $v^g$, starting depot $m_0$
\Statex \textbf{Output:} UGV route $\tau^g = (X^g, T^g) = [(x^g_i,t^g_i)]$

\State UGV navigation waypoints $X^g \gets TSP(\mathcal{M}_r,m_0,v^g)$
\State UGV route starting instance $\tau^g = [ (x^g_0,t^g_0)] $
\For{$x^g_i$ in $X^g$} 
\State $t^g_{i} = t^g_{i-1} + \frac{x^g_{i}-x^g_{i-1}}{v^g} $
\State $\tau^g$.append$(x^g_i,t_i)$
\If{$x^g_i \in \mathcal{M}_r$ }
\State send $t^g_i \rightarrow UAVPlanner $
\State $t^a_i, \mathcal{R}_t \gets UAVPlanner$
\State $t^g_{i} = t^a_{i} + \mathcal{R}_t $
\State $\tau^g$.append$(x^g_i,t^g_i)$
\EndIf
\EndFor
\end{algorithmic}
\end{algorithm}

\subsection{UAV routing}
At the inner level of the proposed framework, we split the full task scenario into subproblems, taking information about the refuel stops provided by the \textbf{\textit{UGVPlanner}}. A task allocation technique is employed to assign distinct assignment points to each subproblem. These subproblems are then individually addressed by formulating them as Energy Constrained Vehicle Routing Problems with Time Windows (E-VRPTW).

\subsubsection{ Allocation of assignment points }

 Given the scenario and the obtained refuel stops $\mathcal{M}_r$ from the MSC algorithm, we can divide the entire problem into $r - 1$ number of subproblems ($r = $ number of refuel stops with starting depot) with an assumption that UGV travels only between two refuel stops in each subproblem. For the subproblem $SP_{i}$, the origin node is refuel stop $m^r_{i-1}$ and the destination node is refuel stop $m^r_{i}$. The subproblems are assigned with separate assignment points. The UAV assignment points covered by the destination refuel stop $m^r_{i}$ are assigned to that subproblem $SP_{i}$. Only, for the first subproblem $SP_{1}$ the assignment points covered by both origin $m^r_{0}$ and destination node $m^r_{1}$ is assigned to it.

Figure \ref{Divison of subproblem} demonstrate the process of subproblem division and task allocation. Figure \ref{entire problem} shows the refuel stops obtained from minimum set cover algorithm which are taken into account for UGV route construction. Based on refuel stops, the first subproblem (figure \ref{subproblem1}) is created by taking the starting depot as the origin node and the refuel stop 1 as the destination node. The UAV assignment points covered by origin node (starting depot) and destination node (refuel stop 1) are assigned for subproblem 1. Similarly, the second subproblem (figure \ref{subproblem2}) is created by taking the refuel stop 1 as origin node and refuel stop 2 as destination node and the assignment points covered by the destination node (refuel stop 2) are assigned for this subproblem. 

Now in the subproblems, the destination nodes $m^r_{i}$ have an \textit{availability} time window constraint because UAV can recharge only when the UGV has already reached the refuel stops. This \textit{availability} time period $t^g_{i}$ is obtained from the \textbf{\textit{UGVPlanner}} and taken in account while modelling the subproblems as energy constrained vehicle routing problem with time windows (E-VRPTW).

\subsubsection{E-VRPTW formulation}
The formulation of the E-VRPTW can be described with a graph theory. Consider an undirected graph $G = (V, E)$ where $V$ is the set of vertices $V = \{S,0,1,2,...D\}$ and  $E$ is the set of edges between the vertices  $i$ and $j$ as $ E= \{(i, j) \, \| \ i, j \, \in \, V, i \neq j \}$. The non-negative arc cost between the vertices  $i$ and $j$ is expressed as $t_{ij}$ and $x_{ij}$ is a binary decision variable whose value will be 1 if a vehicle travels from $i$ to $j$, and 0 otherwise. The UAV will start from refuel stop $S$ and meet the UGV at destination stop $D$.  We then formulated the objective function of the E-VRPTW problem with fuel constraint, time window constraint, optional node constraints as follow:
 
\begingroup
\addtolength{\jot}{0.5em}
\begin{align}
&\min \sum_i \sum_j t_{i j} x_{i j} \quad \forall i, j \in V  \label{eq:1}\\ 
&\sum_{j \in V} x_{i j}=1 \quad \forall i \in V \setminus \{S,D\} \label{eq:2} \\ 
&\sum_{i \in V} x_{i j}=1 \quad \forall j \in V \setminus \{S,D\} \label{eq:3} \\
&\sum_{j \in V} x_{S j}=\sum_{i \in V} x_{i D}=1  \label{eq:6} \\
&f^a_j \leq f^a_i-\left(\mathcal{P}^a(v^a)t_{i j}x_{i j}\right) + L_1\left(1-x_{i j}\right) \quad \forall i, j \in V \setminus \{S,D\} \label{eq:7} \\
&f^a_j=F^a \quad \forall j \in D  \label{eq:8} \\
&0 \leq f^a_j \leq F^a, \quad \forall j \in V   \label{eq:9}\\
&t_j \geq t_i+ (t_{i j} x_{i j})-L_2\left(1-x_{i j}\right) \quad \forall i, j \in V \label{eq:10} \\
&t_{j,start} \leq t_j \leq t_{j,end}, \quad \forall j \in D \label{eq:11} \\
&x_{i j}=0, \quad \forall i \in D, \forall j \in V  \label{eq:16}  \\
&x_{i j} \in\{0,1\}, \quad \forall i, j \in V \label{eq:17} \\
&f_i>0, f_i \in \mathbb{R}{ }_{+} \quad \forall i \in V \label{eq:18} \\
&t_i>0, t_i \in \mathbb{Z} \quad \forall i \in V \label{eq:19} \\
&L_1, L_2>0, \quad L_1, L_2 \in \mathbb{R}{ }_{+} \label{eq:22}
\end{align}
\endgroup

The objective of Eq. \ref{eq:1} is to minimize the total time spent by the UAV. Constraints in Eq. \ref{eq:2} and Eq. \ref{eq:3} represent flow conservation, where the inflow should equal the outflow at any of the assignment point vertices. Following that, constraint in Eq. \ref{eq:6} represents flow conservation for start and end vertices, where the number of UAVs leaving the start vertex must equal the number of UAVs arriving at the end vertex. The Miller-Tucker Zemlin (MTZ) formulation \cite{miller1960integer} for sub-tour elimination is the constraint in Eq. \ref{eq:7}. The MTZ constraint ensures that each node is visited sequentially by keeping track of values such as fuel capacity and power consumption of the UAV corresponding to each node. It ensures that if a node is visited twice, the constraint is broken. This constraint allows  that the UAV’s energy is not fully drained out while eliminating loops. $L_1$ denotes a large number in this constraint. This constraint activates only when there is a flow between vertices $i$ and $j$ and drains the UAV energy based on the time taken between the two vertices. The $\mathcal{P}^a$ represents the UAV's power consumption curve during traversal.

According to constraint Eq. \ref{eq:8}, if the vertex is the destination stop ( recharging stop), the UGV must refuel the UAV to its full capacity $F^a$. Constraint Eq. \ref{eq:9} states that the UAV's fuel should be between $0$ and maximum fuel capacity $F^a$ at any vertex in $V$. The cumulative arrival time at the $j^{th}$ node is equal to the sum of the cumulative time at the node $i$, $t_i$ and the travel time between nodes $i$ and $j$, $t_{ij}$. Here, $L_2$ is a large number that aids in the elimination of sub-tour constraints, as in Eq. \ref{eq:10}. 

Eq. \ref{eq:11} puts a time window constraint that instructs the vehicle to visit the destination node within it's time window, that means the UAV is only allowed to visit the destination node  only when the UGV has reached there. 
The constraint in Eq. \ref{eq:16} indicates that there should be no flow once the vehicle reaches the end node and the route will end there. Eq. \ref{eq:17} is a binary decision variable in charge of flow between the edges. The continuous decision variable, Eq. \ref{eq:18}, monitors the fuel level at any node and has zero as the lower bound value. Eq. \ref{eq:19} denotes the integer decision variable that computes the cumulative time of the UAV's route and has a lower bound of zero. The authors resorted to constrained programming method that provided quality inner-level solutions in a shorter simulation time.

\subsubsection{UAV Planner} 

\begin{algorithm}
\caption{UAV Planner}\label{alg:UAVPlanner}
\begin{algorithmic}[1]
\Statex \textbf{Input:} Set of subproblems   $\mathcal{S} = [SP_i] \gets Task\ Allocation $
\Statex \textbf{Output:} UAV route $\tau^a = (X^a, T^a) = [(x^a_i,t^a_i)]$
\State $\tau^a = [\ ] $
\For{$SP_i$ in $\mathcal{S}$} 
\State $t^g_i \gets UGVPlanner$
\State send $t^g_i \rightarrow SP_i$
\State $(x^a_i,t^a_i), \mathcal{R}_t \gets EVRPTW$
\State $\tau^a$.append$(x^a_i,t^a_i)$
\State send $t^a_i, \mathcal{R}_t  \rightarrow UGVPlanner$
\EndFor
\end{algorithmic}
\end{algorithm}

\begin{figure*}[t]
\centering
\begin{subfigure}[]{0.31\textwidth}
         \includegraphics[ scale=0.4]{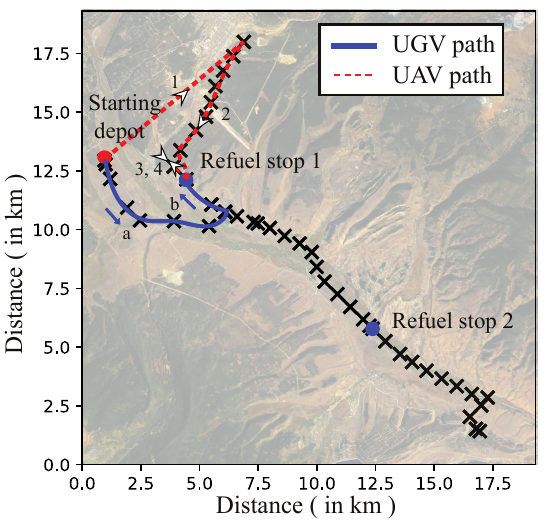 }
         \caption{solution of subproblem 1}  \label{SP1 sol}
\end{subfigure}
\begin{subfigure}[]{0.31\textwidth}
         \includegraphics[ scale=0.4]{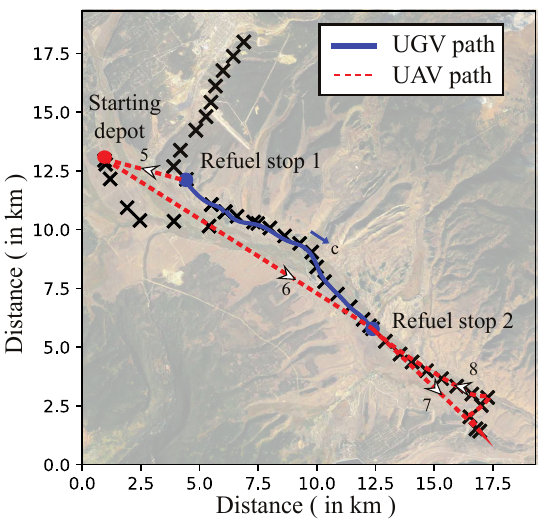}
         \caption{solution of subproblem 2 }
         \label{SP2 sol}
\end{subfigure}
\begin{subfigure}[]{0.31\textwidth}
         \includegraphics[ scale=0.4]{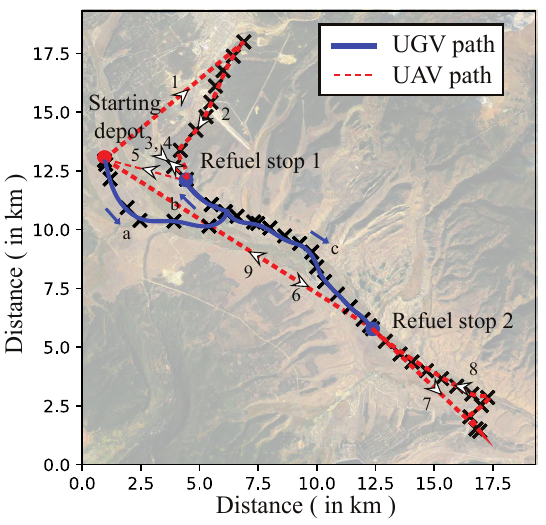}
         \caption{entire solution}
         \label{entire problem sol}
\end{subfigure}

\caption{  a) UAV-UGV routes from subproblem 1 b)UAV-UGV routes from subproblem 2 c) UAV-UGV routes for entire task scenario after combining subproblem 1 \& 2 }
\label{Solution}
\end{figure*}

By solving this E-VRPTW for subproblem $SP_i$, \textbf{\textit{UAVPlanner}} gets the optimal UAV route (both spatial and temporal component) $\tau^a_i = (x^a_i,t^a_i)$ as well as the time instance  $t^a_i$ at which the UAV will arrive at the refuel stop $m^r_{i}$ to recharge with the UGV and the recharging time  $\mathcal{R}_t$  of it, which is dependent on its fuel consumption level. These information are fed back to the \textit{\textbf{UGVPlanner}} again to calculate the UGV \textit{availability} time window for next subproblem $SP_{i+1}$. This reciprocal and iterative process (line 3 - 7 in algorithm \ref{alg:UAVPlanner}) between the \textbf{\textit{UAVPlanner}} and \textbf{\textit{UGVPlanner}} is what facilitates the cooperative route for the entire task scenario. In figure \ref{SP1 sol} and figure \ref{SP2 sol}, we got the routes for the UAV and the UGV which are combined together to get the complete routes of UAV and UGV for the entire task scenario (figure \ref{entire problem sol}).

\section{Results }

We implemented the proposed framework across diverse random task scenarios to evaluate its proficiency. The task scenarios, generated at three distinct scales, helped us investigate the impact of UAV fuel capacity on the overall routing process. In these tests, we compared the results of the greedy and constrained programming methods when applied to the outer-loop baseline of our proposed framework. Additionally, to ascertain the upper limit of the performance metrics, we also constructed a UGV-only route in each scenario, which facilitated the assessment of the practicality and advantages of the cooperative UAV-UGV route in each specific scenario.

\subsection{Design of experiments}
\begin{figure*}[htbp]
\centering
\begin{subfigure}[b]{0.32\textwidth}
         \centering
         \includegraphics[ scale=0.35]{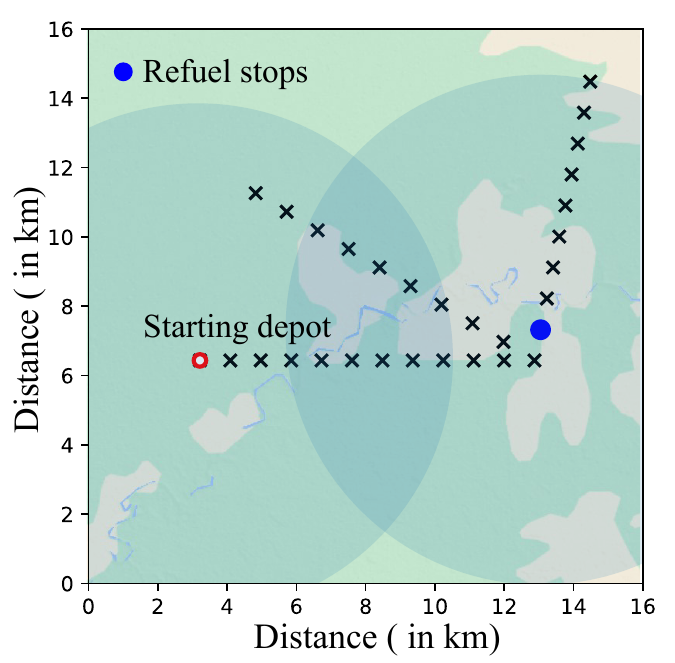}
         \caption{small scale scenario}
         
\end{subfigure}
\hfill     
\begin{subfigure}[b]{0.32\textwidth}
         \centering
         \includegraphics[ scale=0.35]{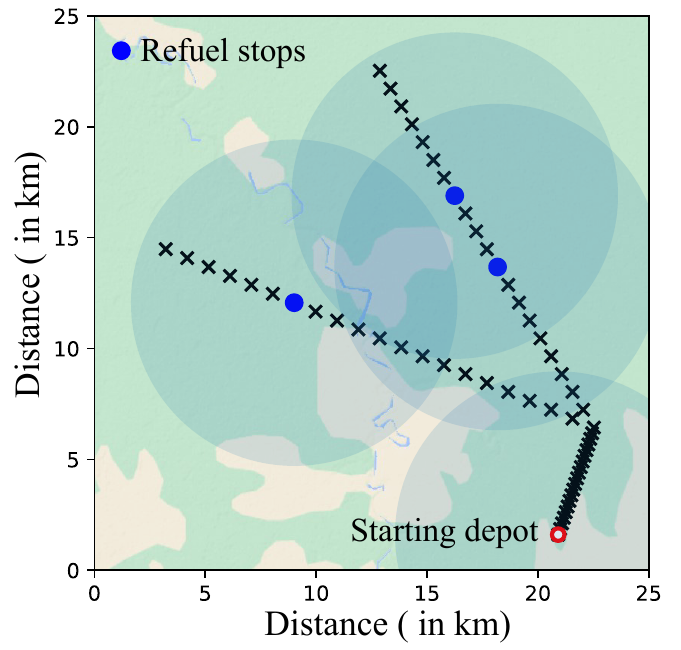}
         \caption{medium scale scenario}
         
\end{subfigure}
\hfill    
\begin{subfigure}[b]{0.32\textwidth}
         \centering
         \includegraphics[ scale=0.35]{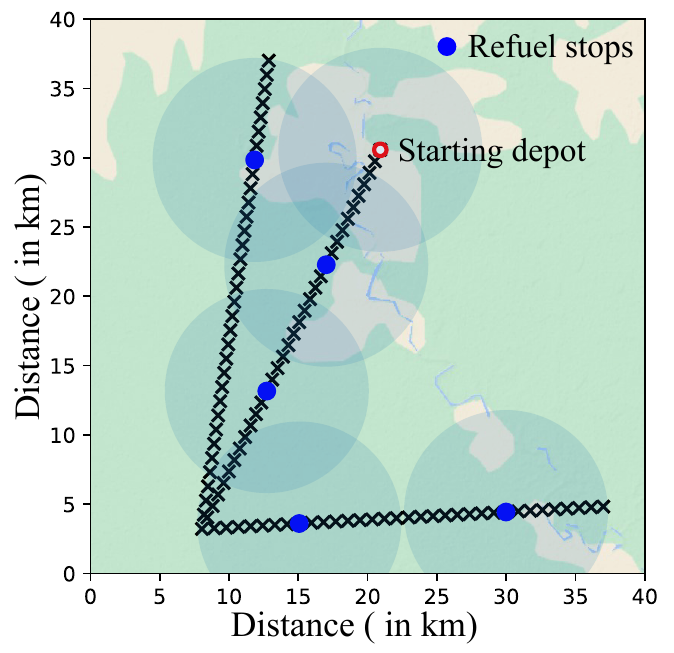}
         \caption{large scale scenario}
         \label{}
\end{subfigure}

\caption{Sample scenarios with starting depot and refuel stops obtained from minimum set cover algorithm.  Radial circles are indicating the coverage area of the UAV from the stops. As the scale of the task scenario grows, the number of task points, number of refueling stops have grown proportionally. }
\label{Scenarios_fig}
\end{figure*}

The efficacy of our proposed framework was tested across numerous random task scenarios generated at three separate scales. We designed the scenarios such that the farthest assignment point from the starting depot was always outside the UAV's radial coverage, guaranteeing that at least one refueling stop was necessary for the UAV to complete the task. To substantiate the robustness and adaptability of the suggested methodology, we experimented with three distinct scales of task instances, as exhibited in Table~\ref{Scenarios}. We introduced a \textit{scale factor}, a non-dimensional number, to represent the relationship between the scenario map size and the UAV's radial fuel coverage area. Three examples of scenarios from three different scales are shown in figure \ref{Scenarios_fig}.

\begin{equation}
\label{SF}
\text{\textit{scale factor}} = \frac{\text{Area of scenario }}{\text{ UAV coverage area on a single charge} }
\end{equation}

\begin{table}[b]
  \centering
  \caption{Specifications of task Scenario}
  \vskip 0.15in
  \begin{small}
  \renewcommand{\arraystretch}{1}
  \begin{tabular}{cccc}
  \hline
   Scale & Map size &  \textit{scale factor} & No. of task points \\

    \hline
   \\Small & 16 km x 16 km &  \centering1.5 &  30 \\
   \\Medium & 25 km x 25 km &  \centering3 &  60\\
   \\Large & 40 km x 40 km &  \centering9 &  100\\
   \\
   \hline
  \end{tabular}
  \label{Scenarios}
  \end{small}
\end{table}

For each instance, two types of cooperative routes (if different) were generated by employing the Greedy method and the CP method at the outer loop of the suggested framework. The UGV-only route (UGV operates alone) was also determined for the specific scenarios. There is no benchmark solution exists to this specific problem due to its complex combinatorial nature, hence we treated the UGV-only route as the baseline method for comparison.   Comparison was made between the cooperative routing route and UGV only route, which signifies the impact of cooperation between UAV and UGV on the task execution. The total task completion time and total energy consumption were treated as the metrics for the evaluation of routes. 
\subsection{Time metrics}

\begin{table*}[]
\centering
\caption{Time metrics of different scenarios}
\resizebox{\linewidth}{!}{%
\begin{tabular}{ccccccc}
\hline
\multirow{3}{*}{\textbf{Map   Size}} & \multirow{3}{*}{\textbf{Scenarios}} & \multicolumn{3}{c}{\multirow{3}{*}{\textbf{Route Time (min.)}}} & \multicolumn{2}{c}{\multirow{3}{*}{\textbf{Improvement (\%)}}} \\
 &  & \multicolumn{3}{c}{} & \multicolumn{2}{c}{} \\
 &  & \multicolumn{3}{c}{} & \multicolumn{2}{c}{} \\ \cline{3-5}
 &  & \multicolumn{2}{c}{\multirow{2}{*}{Cooperative Routing}} & \multirow{2}{*}{UGV only} & \multicolumn{2}{c}{} \\
 &  & \multicolumn{2}{c}{} &  & \multicolumn{2}{c}{} \\ \cline{3-4} \cline{6-7} \\
 &  & Greedy Method & CP Method &  & Greedy Method & CP Method \\ \\  \hline \\
\multirow{10}{*}{Small scale} & Scenario 1 & 200 & 210 & 249 & 19.68 & 15.66 \\
 & Scenario 2 & 91 & 82 & 133 & 31.58 & 38.35 \\
 & Scenario 3 & 117 & 115 & 194 & 39.69 & 40.72 \\
 & Scenario 4 & 153 & 153 & 178 & 14.04 & 14.04 \\
 & Scenario 5 & 148 & 148 & 159 & 6.92 & 6.92 \\
 & Scenario 6 & 222 & 222 & 289 & 23.18 & 23.18 \\
 & Scenario 7 & 190 & 149 & 196 & 3.06 & 23.98 \\
 & Scenario 8 & 128 & 128 & 198 & 35.35 & 35.35 \\
 & Scenario 9 & 222 & 210 & 303 & 26.73 & 30.69 \\
 & Scenario 10 & 223 & 128 & 214 & -4.21 & 40.19 \\ \\ \hline \\
\multirow{10}{*}{Medium Scale} & \multirow{1}{*}{Scenario   1} & \multirow{1}{*}{411} & \multirow{1}{*}{404} & \multirow{1}{*}{497} & \multirow{1}{*}{17.30} & \multirow{1}{*}{18.71} \\
 & Scenario 2 & 414 & 409 & 622 & 33.44 & 34.24 \\
 & Scenario 3 & 364 & 364 & 511 & 28.77 & 28.77 \\
 & Scenario 4 & 340 & 338 & 452 & 24.78 & 25.22 \\
 & Scenario 5 & 297 & 297 & 341 & 12.90 & 12.90 \\
 & Scenario 6 & 396 & 417 & 524 & 24.43 & 20.42 \\
 & Scenario 7 & 296 & 298 & 370 & 20.00 & 19.46 \\
 & Scenario 8 & 324 & 325 & 477 & 32.08 & 31.87 \\
 & Scenario 9 & 292 & 295 & 403 & 27.54 & 26.80 \\
 & Scenario 10 & 291 & 257 & 459 & 36.60 & 44.01 \\ \\ \hline \\
\multirow{10}{*}{Large Scale} & Scenario 1 & 461 & 407 & 438 & -5.25 & 7.08 \\
 & Scenario 2 & 440 & 431 & 467 & 5.78 & 7.71 \\
 & Scenario 3 & 537 & 532 & 484 & -10.95 & -9.92 \\
 & Scenario 4 & 611 & 519 & 462 & -32.25 & -12.34 \\
 & Scenario 5 & 744 & 744 & 680 & -9.41 & -9.41 \\
 & Scenario 6 & 452 & 412 & 436 & -3.67 & 5.50 \\
 & Scenario 7 & 655 & 701 & 607 & -7.91 & -15.49 \\
 & Scenario 8 & 682 & 684 & 588 & -15.99 & -16.33 \\
 & Scenario 9 & 620 & 620 & 570 & -8.77 & -8.77 \\
 & Scenario 10 & 613 & 604 & 537 & -14.15 & -12.48 \\ \\ \hline \\
\end{tabular}}
\label{Time_metric}
\end{table*}

\begin{figure}[htbp]
\centering
\includegraphics[ scale=0.55]{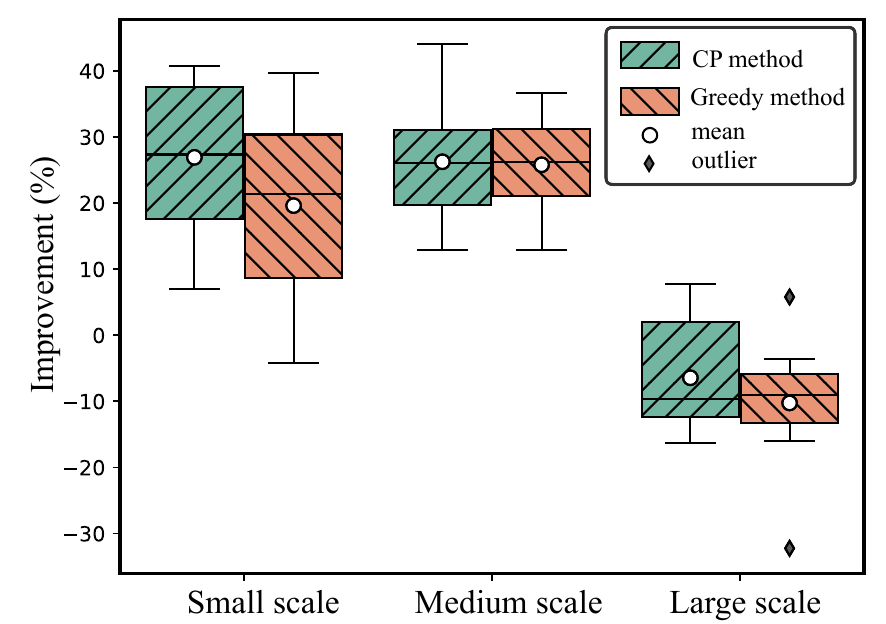}
\caption{Time metrics on 3 different scale of scenarios}
\label{Improvement}
\end{figure}

In Table \ref{Time_metric}, the total task completion time of the route obtained by the three aforementioned methods have been displayed. For all instances in the small-scale scenarios, cooperative routing with the Constrained Programming method in the framework's outer loop proved more time-efficient than the UGV-only routing. The task completion time for UGV-only routes was reduced by approximately 6\% to 40\% through the cooperation between the UAV and UGV in small-scale scenarios. Although cooperative routing with the Greedy method in the outer loop baseline didn't perform as good as the CP method, it was more time-efficient than the UGV-only route in most instances. However, for scenario 10 in small scale, the Greedy method couldn't improve the task completion time through the operative route.

For medium-scale scenarios, the task completion time improved by 12\% up to 45\% through the CP method-based cooperative routing, while for the Greedy method-based cooperative route, the improvement range was 12\% up to 30\%. The cooperative route was more economical than the UGV individual route for most scenarios with the CP method at the outer loop, although the improvement range was a bit less than that in the small-scale scenarios.

However, for most large-scale scenarios, the cooperative route couldn't improve the total task completion time, making the UGV-only route the optimal choice. The figure \ref{Improvement} depicts the total task completion time of the UGV-only route and the respective cooperative routing route with both CP method and Greedy method at the outer loop baseline for three types of scenarios.

\subsection{Energy metrics}
 The energy consumed during the routing process was also analyzed across different scenarios, as demonstrated in Table ~\ref{Energy_m}. The improvement percentage reflects the relative gain in total energy consumption that was achieved through UAV-UGV cooperation. The results confirmed that cooperative routing is more energy-efficient than UGV-only routing. Among the cooperative routing methods, the Constrained Programming (CP) method applied in the outer loop outperformed the Greedy method.

\begin{figure}[htbp]
\centering
\includegraphics[ scale=0.55]{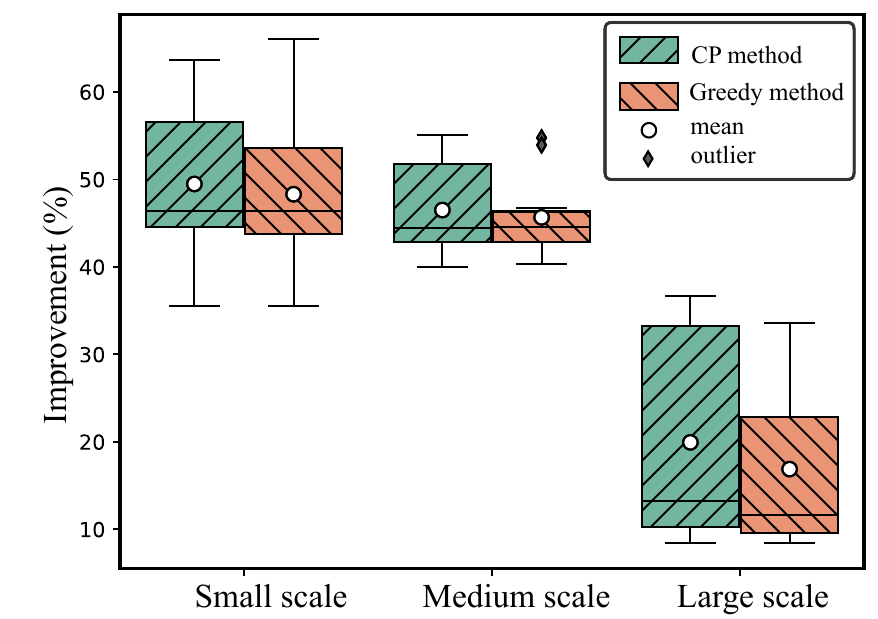}
\caption{Energy metrics on 3 different scale of scenarios}
\label{Energy metrics}
\end{figure}

For the small-scale instances, cooperative routing enabled energy savings ranging from 28\% up to 58\%. For medium-scale scenarios, the improvement ranged from 40-55\%, while for large-scale scenarios, the range was between 8-37\%. This data affirms that cooperative routing, particularly when employing the CP method in the outer loop, can significantly enhance energy efficiency across a variety of scenarios (see figure \ref{Energy metrics}). 

\begin{table*}[]
\centering
\caption{Energy metrics of different scenarios}
\resizebox{\linewidth}{!}{%
\begin{tabular}{ccccccc}
\hline
\multirow{3}{*}{\textbf{Map   Size}} & \multirow{3}{*}{\textbf{Scenarios}} & \multicolumn{3}{c}{\multirow{3}{*}{\textbf{Total energy consumption (MJ)}}} & \multicolumn{2}{c}{\multirow{3}{*}{\textbf{Improvement (\%)}}} \\
 &  & \multicolumn{3}{c}{} & \multicolumn{2}{c}{} \\
 &  & \multicolumn{3}{c}{} & \multicolumn{2}{c}{} \\ \cline{3-5}
 &  & \multicolumn{2}{c}{\multirow{2}{*}{Cooperative Routing}} & \multirow{2}{*}{UGV only} & \multicolumn{2}{c}{} \\
 &  & \multicolumn{2}{c}{} &  & \multicolumn{2}{c}{} \\ \cline{3-4} \cline{6-7} \\
 &  & Greedy Method & CP Method &  & Greedy Method & CP Method \\ \\  \hline \\
\multirow{10}{*}{Small scale} & Scenario 1 & 20.56 & 20.61 & 37.07 & 44.54 & 44.40 \\
 & Scenario 2 & 6.73 & 7.19 & 19.80 & 66.00 & 63.68 \\
 & Scenario 3 & 12.53 & 11.82 & 28.88 & 56.62 & 59.08 \\
 & Scenario 4 & 17.08 & 17.08 & 26.50 & 35.56 & 35.56 \\
 & Scenario 5 & 12.46 & 12.46 & 23.67 & 47.38 & 47.38 \\
 & Scenario 6 & 23.47 & 23.47 & 43.03 & 45.44 & 45.44 \\
 & Scenario 7 & 14.14 & 16.72 & 29.18 & 51.56 & 42.71 \\
 & Scenario 8 & 13.49 & 13.49 & 29.48 & 54.25 & 54.25 \\
 & Scenario 9 & 25.47 & 24.86 & 45.11 & 43.55 & 44.89 \\
 & Scenario 10 & 19.73 & 13.60 & 31.86 & 38.07 & 57.32 \\ \\ \hline \\
\multirow{10}{*}{Medium Scale} & \multirow{1}{*}{Scenario   1} & \multirow{1}{*}{39.40} & \multirow{1}{*}{34.00} & \multirow{1}{*}{73.99} & \multirow{1}{*}{46.75} & \multirow{1}{*}{54.04} \\
 & Scenario 2 & 41.90 & 41.63 & 92.60 & 54.75 & 55.04 \\
 & Scenario 3 & 41.82 & 41.82 & 76.08 & 45.03 & 45.03 \\
 & Scenario 4 & 38.15 & 38.05 & 67.29 & 43.31 & 43.46 \\
 & Scenario 5 & 30.10 & 30.10 & 50.77 & 40.71 & 40.72 \\
 & Scenario 6 & 46.52 & 46.85 & 78.01 & 40.37 & 39.94 \\
 & Scenario 7 & 31.56 & 31.57 & 55.09 & 42.71 & 42.70 \\
 & Scenario 8 & 39.75 & 39.67 & 71.02 & 44.03 & 44.15 \\
 & Scenario 9 & 33.01 & 33.14 & 60.00 & 44.98 & 44.76 \\
 & Scenario 10 & 31.49 & 30.71 & 68.34 & 53.92 & 55.06 \\  \\ \hline \\
\multirow{10}{*}{Large Scale}  & Scenario 1 & 43.63 & 41.27 & 65.21 & 33.09 & 36.71 \\
 & Scenario 2 & 46.17 & 45.35 & 69.53 & 33.59 & 34.78 \\
 & Scenario 3 & 62.94 & 62.67 & 72.06 & 12.66 & 13.03 \\
 & Scenario 4 & 56.05 & 59.50 & 68.78 & 18.51 & 13.50 \\
 & Scenario 5 & 92.74 & 92.74 & 101.24 & 8.39 & 8.39 \\
 & Scenario 6 & 49.15 & 42.32 & 64.91 & 24.28 & 34.81 \\
 & Scenario 7 & 81.63 & 64.41 & 90.37 & 9.68 & 28.73 \\
 & Scenario 8 & 80.17 & 79.96 & 87.54 & 8.42 & 8.66 \\
 & Scenario 9 & 75.80 & 75.80 & 84.86 & 10.67 & 10.67 \\
 & Scenario 10 & 72.37 & 71.89 & 79.95 & 9.48 & 10.07 \\  \\ \hline \\
\end{tabular}}
\label{Energy_m}
\end{table*}

 \subsection{Computational time}
\begin{figure}[htbp]
\centering
\includegraphics[ scale=0.55]{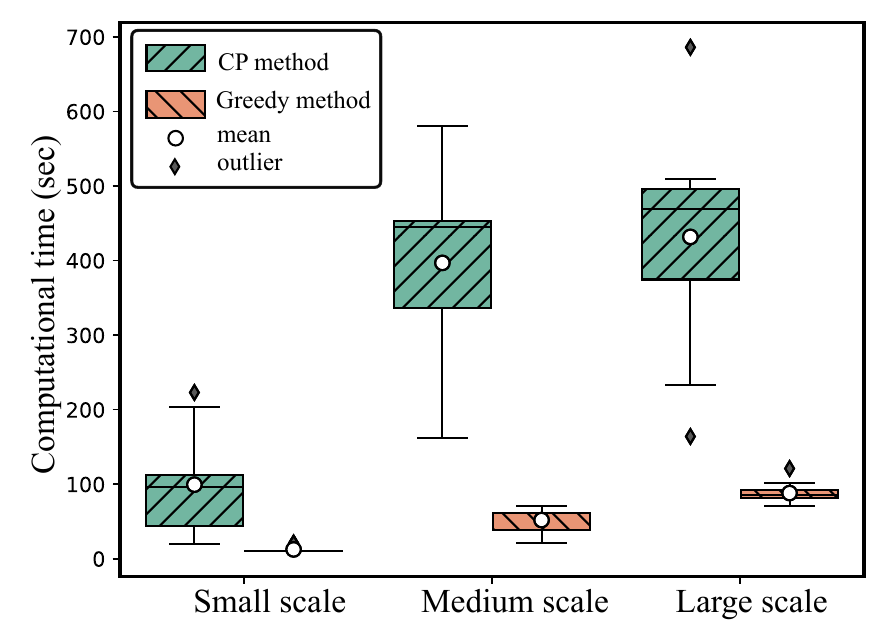}
\caption{Computational time}
\label{computational time}
\end{figure}

 For real-time applications, the computational time of the vehicle routing problem is a crucial factor. The greedy method and the Constrained Programming (CP) method, when implemented at the outer loop of the proposed framework, display notable differences in computational time. As shown in figure \ref{computational time}, the greedy method requires substantially less computational time compared to the CP method.

Given the subproblem division approach at the inner loop, the computational time increases for both the Greedy and CP methods as the scale of the scenarios increases. However, the Greedy method consistently outperforms the CP method, and the gap between their respective computational times grows in proportion to the scale of the scenario. This highlights the Greedy method's efficiency and its suitability for larger scale scenarios requiring rapid computations.

\section{Discussion}

The geometric configuration of assignment target points within a scenario critically influences the potential for improving total task completion time through cooperative routing. As observed in figure \ref{Improvement}, there are positive improvement percentages for small and medium scale scenarios, but a negative trend emerges for large scale scenarios. The total time taken to complete the task via a cooperative route depends on three elements: UAV traversal time, UGV traversal time, and the waiting time of the UGV during UAV refueling. Conversely, the total task completion time for the UGV-alone route depends solely on the UGV traversal time, as it does not involve refueling.

\begin{figure}[htbp]
\centering
\includegraphics[ scale=0.35]{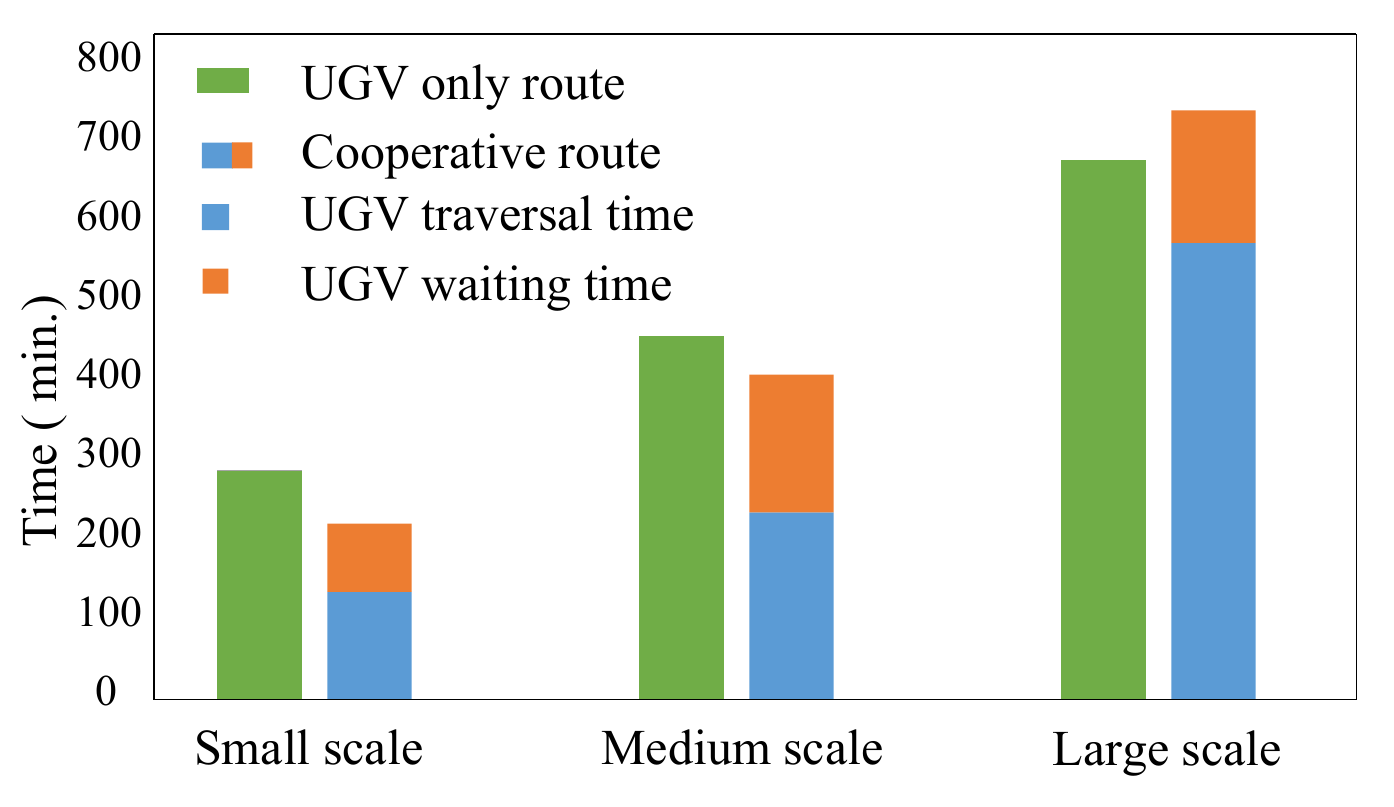}
\caption{Cooperative routing Vs UGV-only route}
\label{UGV_only vs copperation}
\end{figure}

When the UGV operates alone, it has to visit all assignment target points alone, leading to a significant UGV traversal time. This duration increases proportionally with the scale of the scenario, as demonstrated in figure \ref{UGV_only vs copperation}. However, this UGV traversal time can be reduced through cooperative routing, which divides the target points between the UAV and UGV. However, a drawback of cooperative routing is the addition of waiting time during which the UAV is refueled by the UGV.

In small and medium scale scenarios, the asignment points spread is limited and therefore there are fewer refueling stops. This means the extra waiting time at refueling stops never exceeds the reduction in UGV traversal time, leading to a shorter total task completion time for the cooperative route compared to the UGV-alone route. However, in large scale scenarios, where there are many refueling stops due to the extensive assignment points spread, the additional waiting time can surpass the decrease in UGV traversal time. This results in the cooperative route taking longer overall than the UGV-alone route.

In terms of energy consumption metrics, the cooperative route consistently outperforms the UGV-alone route, regardless of the scenario scale. This is because, as demonstrated in figure \ref{Power_curve}, the UAV consumes five times less energy than the UGV per unit distance traveled, making the UGV the dominant influence on total energy consumption. Hence, the UGV-alone route, which involves a longer UGV traversal distance, consumes more energy than the cooperative route, where the UGV covers a smaller distance due to task division with the UAV.

Nevertheless, as the scale of scenarios increases, the gap in total energy consumption between the UGV-alone route and the cooperative route narrows. This is because in larger scenarios, despite cooperation with the UAV, the UGV must still cover a considerable distance to provide suitable refueling stops, leading to higher overall energy consumption.

\section{Conclusion}

In this work, we focus on a cooperative vehicle routing problem involving a UAV-UGV team with fuel constraints. Both vehicles are required to cover a set of assigned task points, with the UAV periodically recharging from the UGV to complete the assignment in minimum possible time. Finding the optimal recharging rendezvous points, in terms of both location and timing, between the UAV and UGV is integral to achieve an optimal route in this cooperative routing problem. We introduce a sequential optimization framework that operates in two primary steps. The initial step involves the utilization of a minimum set cover algorithm to determine the locations for refueling stops. These identified locations serve as an input to the \textit{\textbf{UGVPlanner}}, which then creates the UGV route employing a Traveling Salesman Problem model. In the subsequent step, a task allocation technique is employed to partition the entire problem into smaller, more manageable subproblems. The \textit{\textbf{UAVPlanner}} then develops the UAV route by framing these subproblems as instances of the Energy-Constrained Vehicle Routing Problem with Time Windows (E-VRPTW).

Our framework has been successfully applied to 30 distinct task scenarios across three different scales, showcasing its effectiveness and practicality. The cooperative routes resulting from our framework were benchmarked against the UGV-only routes for the same scenarios which served as an upper limit for comparison. The results reveal substantial improvements, with time consumption reduced by 10-30\% and energy consumption diminished by 15-50\% in most instances through the cooperative routing. In the future direction of the work, we would be expanding the framework for persistent surveillance on the task points and consider the stochasticity in the scenarios. Insights from this study suggest a potential enhancement of leveraging the UGV's idle waiting times during refueling to establish a mobile recharging rendezvous, which will be a focal point in our subsequent investigations.  

\section*{Acknowledgments}
The authors would like to express their gratitude to the DEVCOM Army Research Laboratory for their financial support in funding the projects under grant W911NF-14-S-003.

\section*{Declarations}
\begin{itemize}
\item \textbf{Funding} The work is funded by Army Research Laboratory grant W911NF-14-S-003
\item \textbf{Conflict of Interests} The authors declare no conflict of interest
\item \textbf{Ethics Approval} Not applicable
\item \textbf{Consent to Participate} Not applicable
\item \textbf{Consent for Publication} All authors consent to publication
\item \textbf{Code or data availability} The code and data backing this study's outcomes aren't publicly shared, however it can be obtained upon a credible request. 
\item \textbf{Authors’ contributions} The conceptualization of ideas and methodologies was carried out by MSM, SR, and PAB. Implementation was undertaken by MSM and SR. The initial draft was penned by MSM, while the review and editing process involved JH, JPR, JD, CM, and PAB. The project was overseen and supervised by JH, JPR, JD, CM, and PAB, with PAB also handling project administration. All contributing authors have reviewed and given their consent for the final version of the manuscript to be published.

\end{itemize}

\bibliography{sn-bibliography}


\end{document}